\documentclass[sigconf]{acmart}

\setcopyright{none}
\acmDOI{}
\acmISBN{}

\usepackage{booktabs}
\usepackage{latexsym}
\usepackage{colortbl,array,xcolor}
\usepackage{multirow}
\usepackage{multicol}
\usepackage{xspace}
\usepackage{amsfonts}
\usepackage{tcolorbox}
\tcbuselibrary{breakable}
\usepackage{tikz}
\usepackage{amsmath}
\usetikzlibrary{tikzmark}
\usepackage{utfsym}
\usepackage{microtype}
\usepackage[ruled,vlined]{algorithm2e}
\usepackage{listings}
\usepackage{graphicx}

\makeatletter
\newcommand*\myfontsize{%
  \@setfontsize\myfontsize{6.7}{8}%
}
\makeatother

\definecolor{c1}{HTML}{688990}
\definecolor{myred}{HTML}{C3375A}
\definecolor{mypurple}{HTML}{CD82F2}
\definecolor{mypink}{HTML}{EC9FA8}
\definecolor{cadmiumgreen}{rgb}{0.0, 0.42, 0.24}
\definecolor{myblue}{rgb}{0.2, 0.3, 0.6}

\newcommand{\mytextbox}[2]{\tikzmarknode[draw=#1,thick,inner sep=2pt]{test}{\myfontsize #2}}
\newcommand{\etc}{\textit{etc}\xspace}

\newcommand{\rret}{\textcolor{myred}{\small{\texttt{[Relation Retrieval]}}}}
\newcommand{\ret}{\mytextbox{c1}{\textbf{\textcolor{c1}{\textit{Retrieval Token}}}}}
\newcommand{\eret}{\textcolor{myred}{\small{\texttt{[Entity Retrieval]}}}}

\newcommand{\mgen}{$\mathcal{M}$\xspace}

\newcommand{\mret}{$\mathcal{R}$\xspace}
\newcommand{\model}{\textsc{ArG}\xspace}

\newcommand{\eg}{\emph{e.g.,}\xspace}
\newcommand{\wo}{\emph{w/o.}\xspace}
\newcommand{\crel}{\mytextbox{myblue}{\textbf{\textcolor{myblue}{\textit{Relevance Token}}}}}
\newcommand{\cre}{\mytextbox{mypink}{\textbf{\textcolor{mypink}{\textit{Rationality Token}}}}}
\newcommand{\cuse}{\mytextbox{mypurple}{\textbf{\textcolor{mypurple}{\textit{Utility Token}}}}}
\newcommand{\paratitle}[1]{\noindent\textbf{#1}}

\SetCommentSty{mycommfont}

\title{Learning to Retrieve and Reason on Knowledge Graph through Active Self-Reflection}

\author{Han Zhang}
\email{hanzhang0816@ruc.edu.cn}
\affiliation{%
  \department{School of Statistics\\Center for Applied Statistics}
  \institution{Renmin University of China}
  \city{Beijing}
  \country{China}
}

\author{Langshi Zhou}
\email{lszhou@uw.edu}
\affiliation{%
  \department{Department of Statistics}
  \institution{University of Washington}
  \city{Seattle}
  \state{Washington}
  \country{USA}
}

\author{Hanfang Yang}
\authornote{Corresponding author.}
\email{hyang@ruc.edu.cn}
\affiliation{%
  \department{School of Statistics\\Center for Applied Statistics}
  \institution{Renmin University of China}
  \city{Beijing}
  \country{China}
}

\acmConference[LASS '26]{The 2nd Workshop on LLM Agents for Social Simulation (LASS), co-located with CIKM 2026}{November 8, 2026}{Rome, Italy}
\acmYear{2026}


\begin{document}
\begin{abstract}
Extensive research has investigated the integration of large language models (LLMs) with knowledge graphs to enhance the reasoning process. However, understanding how models perform reasoning utilizing structured graph knowledge remains underexplored. Most existing approaches rely on LLMs or retrievers to make binary judgments regarding the utilization of knowledge, which is too coarse. Meanwhile, there is still a lack of feedback mechanisms for reflection and control throughout the entire reasoning path. This paper proposes an \underline{A}ctive self-\underline{R}eflection framework for knowledge \underline{G}raph reasoning (\model), introducing for the first time an end-to-end training approach to achieve iterative reasoning grounded on structured graphs. Within the framework, the model leverages special tokens to \textbf{\textit{actively}} control the reasoning process, achieving on-demand retrieval and multi-dimensional, finer-grained knowledge evaluation. The reasoning paths generated by the model exhibit high interpretability, enabling deeper exploration of the model's understanding of structured knowledge. Ultimately, the proposed model achieves outstanding results compared to existing baselines in knowledge graph reasoning tasks.
\end{abstract}
\maketitle

\section{Introduction}

Knowledge graphs (KGs), offering structured, explicit, and interconnected knowledge representation, serve as a highly promising external knowledge source for factual reasoning. As a classical reasoning task, knowledge graph question answering (KGQA), remains a significant challenge to efficiently understand graphs. Traditional KGQA approaches can generally be categorized into two main types: Information Retrieval (IR)-based methods and Semantic Parsing (SP)-based methods. Specifically, IR-based methods enhance the generation process by retrieving related entities, relations, triplets to construct a subgraph from which answers are derived. On the other hand, SP-based methods generate logical forms (\eg S-expression~\cite{GrailQA}, SPARQL~\cite{SPARQL}, \etc) that directly interact with KGs to obtain precise answers. 
\begin{figure}[t]
    \centering
    \includegraphics[width=\linewidth]{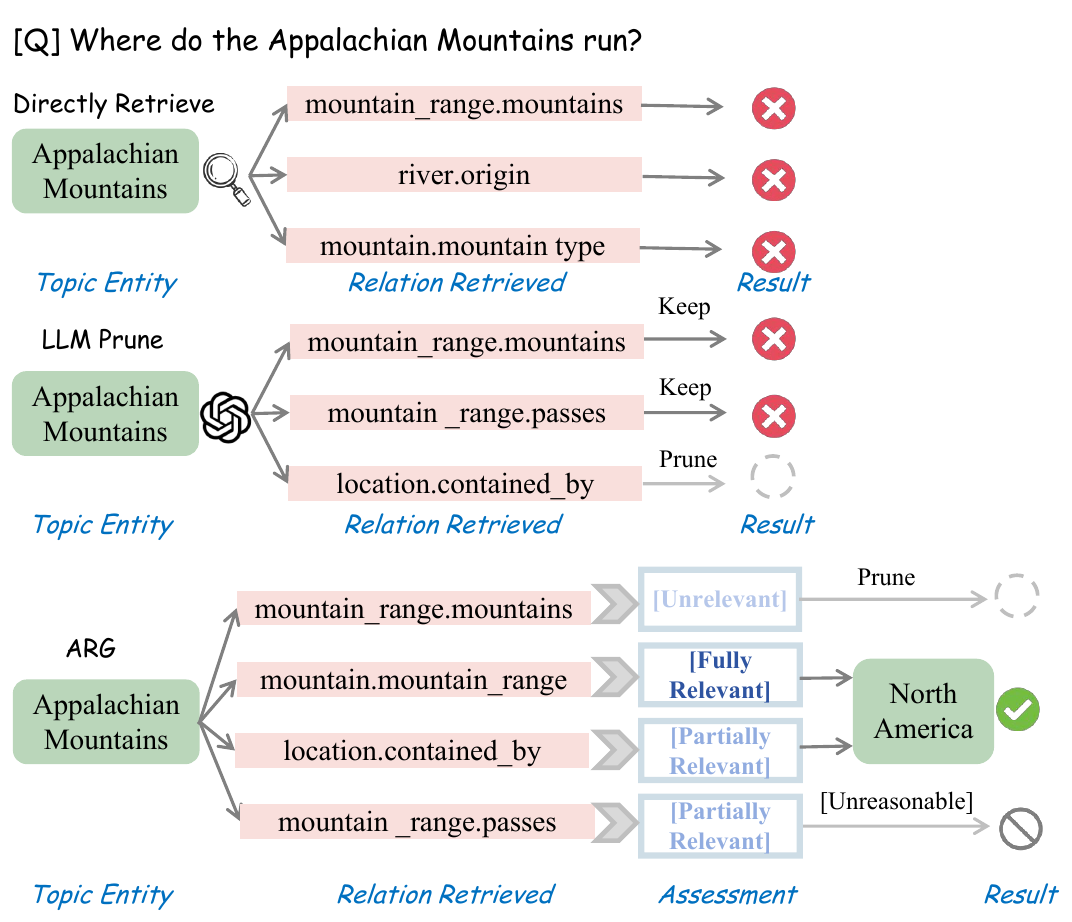}
    \caption{An example of \model performs a more fine-grained assessment and actively retrieves knowledge compared to LLM pruning and direct retrieval. \model not only provides reasonable evaluations, but also captures useful indirect relationships.}
    \label{fig:compare}
\end{figure}
The aforementioned methods both face notable limitations. Although SP-based methods can generate accurate and robust results, they require the annotation of high-quality, and often expensive, logical forms as supervision~\cite{zhang-etal-2022-subgraph}. Furthermore, these methods struggle to capture the procedural information underlying the comprehension of structured graphs, rendering the workflow overly "black-box". IR-based methods typically adopt a \emph{retrieve and generate} framework. However, due to the disparity between natural language and structured graph knowledge, traditional approaches require dedicated training and exhibit low retrieval efficiency~\cite{BM25,Contriever,li-etal-2020-efficient}. 

Recently, there have been some methods combining large language models (LLMs), which have broadened the boundaries of solutions. They leveraged the thinking capabilities of LLMs to derive answers by retrieving facts in a step-wise and training-free manner~\cite{jiang-etal-2023-structgpt,gu-etal-2023-dont,sun2023think}. 

In consideration of all existing methodologies, we raise three practical yet remain underexplored issues: (\romannumeral1) \textbf{Lack of fine-grained relevance assessment.} Existing methods either directly retrieve knowledge employing sparse or dense retrievers or depend on LLMs for binary pruning. However, these methods exhibit limited flexibility, as they fail to offer a fine-grained assessment for determining the extent to which the retrieved knowledge is useful, as depicted in Figure~\ref{fig:compare}. (\romannumeral2) \textbf{Low retrieval effectiveness.} The representation of relations in KGs, such as Freebase~\cite{Freebase}, differs from natural language questions. At the same time, numerous candidates with high lexical similarity exist. Conventional retrieval methods exhibit low effectiveness and often overlook implicit relationships within the structured graph~\cite{liu-etal-2024-knowledge-graph}. (\romannumeral3) \textbf{Lack of rational self-reflection.} During iterative reasoning, it is crucial to terminate exploration of unreasonable paths in a timely manner. Evaluating the coherence of intermediate reasoning steps and the reliability of the final answer is essential for improving the model’s step-by-step reasoning over structured graphs.

Inspired by the rapid advancements in LLM for Retrieval-Augmented Generation (RAG) \cite{asai2023selfrag, zhang-etal-2024-onegen}, this study introduces the \model framework, which integrates specialized \emph{self-reflection tokens} to enhance reasoning capabilities when interacting with structured graph data. By utilizing reasoning paths within the graph as weak supervision signals, the model is trained end-to-end to enable on-demand retrieval and reflective reasoning over knowledge graphs. At each step of the iterative reasoning process, the model determines whether retrieval is necessary. If so, it provides fine-grained relevance evaluation of the retrieved knowledge to the query (e.g., relations and entities) and assigns a rationality score based on the current hop. Moreover, we designed a module called the hypo-generator to enhance retrieval accuracy. The iteration terminates when there is sufficient information to answer the query or an unreasonable path is encountered. The final answer is derived then accompanied by an evaluation of the answer's utility. The reasoning process of \model unfolds as a reasoning tree. Specifically, the candidates generated at each step form the nodes of the tree, enabling parallel expansion of downstream paths. Ultimately, the final score for each leaf node is computed based on backtracking the scores along its path.



In summary, our primary contributions include: (\romannumeral1) We propose an end-to-end training framework that enables the model to perform iterative reasoning over structured graph. (\romannumeral2) We innovatively introduce special tokens in graph reasoning tasks, equipping the model with the capability to perform fine-grained evaluations of the retrieved knowledge and self-reflect on its reasoning process. (\romannumeral3) We introduce a hypo-generator during the inference process to enhance retrieval effectiveness.





\section{Related Work}
\paratitle{Knowledge Graph Question Answering.}
Traditional methods represent entities and relations within an embedding space, leveraging specifically designed model architectures such as Key-Value Memory Networks\cite{miller-etal-2016-key,das-etal-2017-question}, as well as seq2seq frameworks like LSTM-based~\cite{sun-etal-2018-open} and T5-based~\cite{shu-etal-2022-tiara} networks. Recently, leveraging the powerful reasoning capabilities of LLMs, diverse methods have emerged for knowledge graph question answering. Prompt-based approaches, such as KB-BINDER~\cite{kb-binder}, uses few-shot examples to guide the model in generating credible logical forms. Graph-CoT \cite{jin-etal-2024-graph} incorporates Chain-of-Thought (CoT) reasoning to encourage multi-step reasoning over KGs. ToG \cite{sun2023think} iteratively explores related entities and relations on KGs based on LLM-driven reasoning.

Other methods, such as \citet{jiang-etal-2023-structgpt} and \citet{xiong-etal-2024-interactive}, enable iterative KG-based operations by integrating pre-defined functions as an interaction interface. Similarly, \citet{jiang2024kg} treat LLMs as intelligent agents, generating instruction data through KG reasoning programs for fine-tuning the base LLM. In contrast to these prompt-based approaches, we employ an end-to-end training framework, wherein distinct instruction tasks are explicitly defined during the data collection phase while simultaneously enabling procedural reasoning during the inference stage. 


\begin{figure*}[t]
    \centering
    \includegraphics[width=\linewidth]{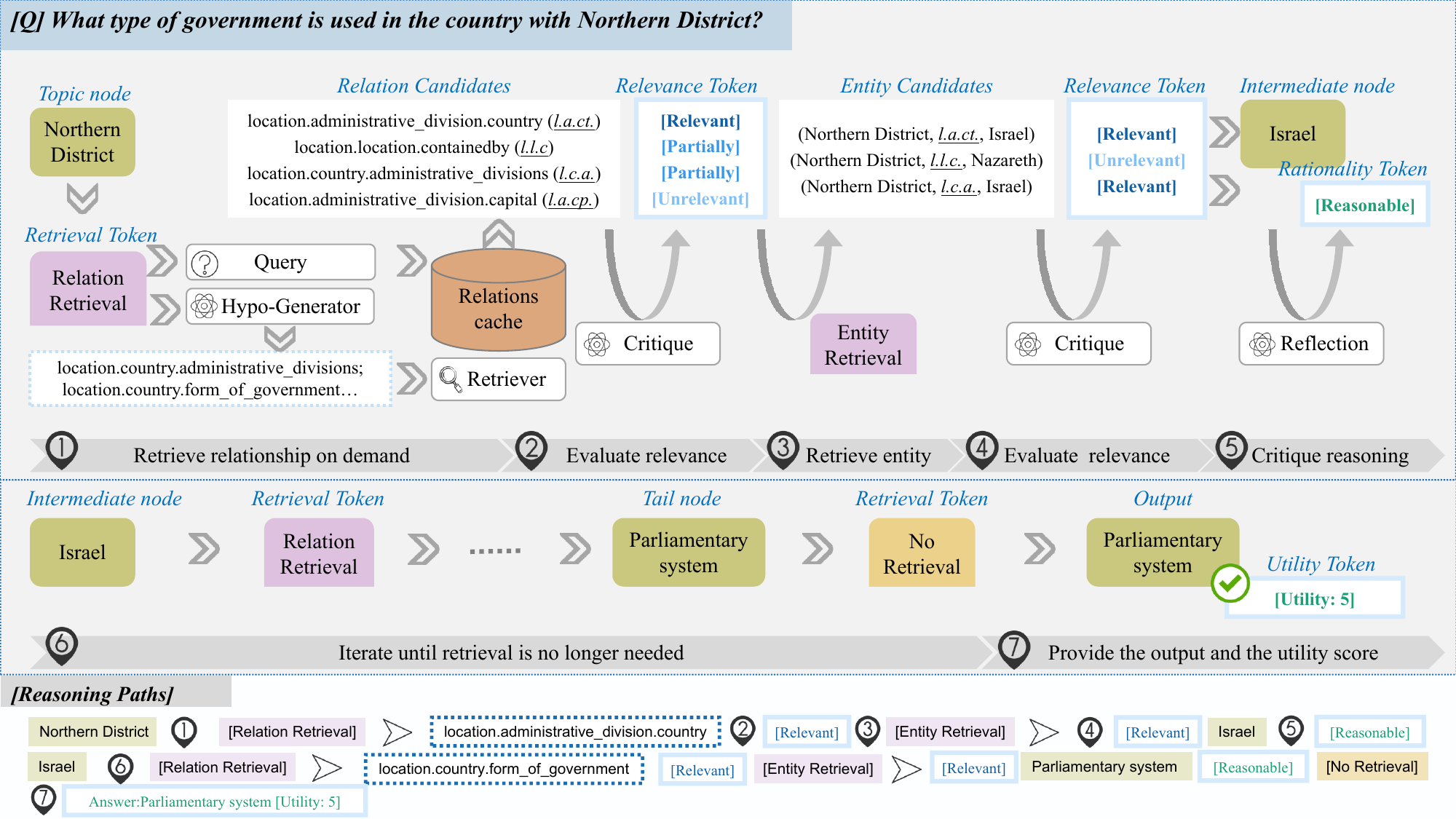}
    \caption{The overall framework of \model. Given an input query, the trained generator model \mgen iteratively performs knowledge retrieval over the structural graph based on the retrieval token. Subsequently, the retrieved knowledge undergoes processes of critique and reflection, where implausible information is filtered. The iterative procedure culminates in the generation of an answer. \model exhibits strong interpretability when applied to structured graph. As demonstrated in the example, the step-by-step reasoning path is organized in the lower half.}
    \label{fig:Self-graph}
\end{figure*}
\paratitle{RAG within Knowledge Graph.} Retrieval-Augmented Generation combines retrieved external knowledge with LLMs for improved task performance, incorporating domain-specific information to ensure factuality and credibility~\cite{guu2020retrieval}.
FLARE~\cite{jiang-etal-2023-active} predicts upcoming sentences to adaptively retrieve relevant passages. Self-RAG~\cite{asai2023selfrag} learns to retrive on-demand guided by reflection tokens. OneGen further unifies retrieval and generation in one model. In the context of knowledge graphs, retrieval focuses primarily on graph databases rather than text corpora, necessitating additional consideration of relationships between texts and the structural information inside. GNN-RAG~\cite{mavromatis2024gnn} integrate Graph Neural Networks (GNNs) with RAG to retrieve reasoning paths. TIARA~\cite{shu-etal-2022-tiara} and ChatKBQA~\cite{luo-etal-2024-chatkbqa} enhance the generation of reliable logical forms by retrieving entities and relations from knowledge graphs. HyKGE~\cite{jiang2024hykge} leverages LLMs to explore feasible directions within medical knowledge graphs. Furthermore, \citet{ji-etal-2024-retrieval} advances the retrieval capabilities via CoT reasoning.

\section{Preliminary}
In this section, we introduce the fundamental concepts of the KGQA task and the definition of reasoning paths.

\textbf{KGQA.} The task of KGQA requires predicting the correct answers based on reasoning over both free-form text $q$ and an inherently structured graph $\mathcal{G}$. The KG here consists of a set of triplets $(s, r, o)$ defined in the RDF format where $s$ is an entity, $r$ is the corresponding relation, and $o$ can be either an entity or a literal value. Each entity $e$ is uniquely identified by a MID and some have a friendly name, \eg $e$.id = \textit{"m.03\_r3"} with its friendly name \textit{"Jamaica"}. Relations in the Freebase KG are defined hierarchically, \eg $r$ = \textit{"location.country.languages\_spoken"}. We assume that the topic entities $\{e_q\}$ and the candidate answers $\{a\}$ are linked to the entities or values in $\mathcal{G}$.

\textbf{Reasoning Path.} We define a valid reasoning path $w$ of depth $D$ is a sequence of connected triplets connecting the topic entity $e_q$ and the answer $a$, which can be also referred to as a $D$ hop path:
\begin{equation}
        w_{1:D}= (e_q, r_1,e_1,\cdots,r_D, a).
\end{equation}
Based on the iterative reasoning process, the objective of the KGQA task can be formulated as the optimization of the following function:
\begin{equation}
P_{\theta}(a|q, \mathcal{G}) = \mathbb{E}_{w_{1:D} \sim Q(w)}\prod_{i=1}^{D} P_{\theta}(w_i|w_{<i}, q, \mathcal{G}),
\end{equation}
where $Q(w)$ denotes the posterior distribution of the faithful reasoning path grounded in the KG.
\begin{table*}[t]
\centering
\footnotesize
\begin{tabular}{lcc}\toprule
Token Type & Category & Definitions \\ \midrule
\textit{Retrieval}    & \{relation, entity, no retrieval\} & trigger knowledge retrieval \\
\textit{Relevance}  &\{relevant, partially, unrelevant\}  &   assess relevance with $q$ \\
\textit{Rationality}  &\{reasonable, partially, unreasonable\}  & evaluate logical coherence   \\
\textit{Utility} &\{5, 4, 3, 2, 1\}  & $a$ is a useful response to $q$.   \\
\bottomrule
\end{tabular}
    \caption{Four types of self-reflection tokens used in \model. The exact token values are summarized in Appendix~\ref{app:tokens}.}
   \label{tab:token} 
\end{table*}
\section{Methodology}
We propose \underline{A}ctive Self-\underline{R}eflection \underline{G}raph Reasoning (\model), the overview of \model at inference is shown in Figure~\ref{fig:Self-graph} and detailed in Algorithm~\ref{alg:workflow}. \model facilitates four types of \emph{self-reflection} tokens (see Table~\ref{tab:token}) to achieve controlled retrieval and evaluation. These tokens are added to the vocabulary during training. The trained language model actively decides whether knowledge retrieval is necessary to answer the query, while correspondingly generating relevance assessments and rationality reflections throughout the graph reasoning process. Irrelevant and unreasonable reasoning will be prevented for further exploration. To further optimize retrieval effectiveness during the retrieval stage, we employ the hypo-generator module, where the trained model is additionally tasked with generating potential relation candidates to serve as supplementary inputs for retrieval. Ultimately, we will form a clear reasoning tree, with each path's score being aggregated from multiple dimensions to arrive at the final answer.

\subsection{\model Training}
\label{sec:train}
To emulate the inferential process within a graph structure, we construct the training dataset based on the reasoning paths. Reasoning paths~\cite{wang2021relational, luo2023reasoning} are capable of capturing rich semantic information between entities. Each reasoning path can be regarded as a logically coherent walk over the knowledge graph.

Directly obtaining labeled inferential paths $\mathcal{W}^*$ from the knowledge graph often proves to be challenging. Instead, we can typically leverage graph search algorithms to identify the shortest paths connecting the topic entity and the target candidates. While these paths may not always correspond to the optimal reasoning paths, they are grounded in the knowledge graph and offer valuable insights toward deriving the answer, as shown in the following example.

\textbf{Example.} Given the question \emph{"Which countries border the US"}, one of the labeled reasoning paths is: $w$ = \emph{US} $\xrightarrow{\texttt{adjoin}}$ \emph{Canada}. At the same time, another valid reasoning path exists: $w^{\prime}$ = \emph{US} $\xrightarrow{\texttt{contains}} $\emph{Columbia River} $\xrightarrow{\texttt{flow\_through}}$ \emph{Canada}.

Although the latter path $w^{\prime}$ does not directly provide the adjacency information between the two countries, it uses the fact that the \emph{Columbia River} flows through \emph{Canada} and arrives at the correct entity. Consequently, we leverage the reasoning paths grounded within the graph and design curated self-reflection tasks to insert reflection tokens, ultimately training the generator model.

\paragraph{Weakly Supervised Data Collecter.}

We summarize the process of data collection in Algorithm~\ref{algo:data_gen}. We extract the shortest paths connecting the questions and answers as supervisory reasoning paths $w$, following RoG~\cite{luo2023reasoning}. For each reasoning path, we augment it with candidate sets for both the relations and entities involved. A retriever $\mathcal{R}$ is employed to retrieve $K$ semantically relevant relations as candidate relations $\mathbf{C_r}$. Similarly, all tail nodes at the current hop are retrieved as candidate entities $\mathbf{C_e}$.
\begin{algorithm}
\small
\caption{\model Workflow}
\label{alg:workflow}
\SetKwRepeat{Do}{do}{while}
\SetKwInOut{Input}{Input}
\SetKwInOut{Output}{Output}
 {\bfseries Require: }{Generator model \mgen, Retriever \mret, KG $\mathcal{G}$}\\
 {\bfseries Input: }{Topic entity $e_q$ and input query $q$}\\
 {\bfseries Output: }{predicted prediction tree $T$.}\\
 {\bfseries Initialization: } $T = [w_0], d=1, w_0=(e_q,)$ \\
\While {$d$ $\leq$ $D_{max}$} {
\mgen predicts \ret~given $(q,w_{<d})$\;
\uIf{\ret~==\textcolor{myred}{\small{\texttt{[Relation Retrieval]}}}}{
$\widehat{r_d}$ = $\mathcal{M}(q, w_{<d})$ \tcp*{\textcolor{blue}{Hypo-Generator}}
Retrieve relevant relationship $\mathbf{R}$ using \mret given $(q,e_{d-1}, \widehat{r_d})$ \;
\ForEach{$r_d \in \mathbf{R}$ }{\mgen predicts \crel \;
 \uIf{\crel $\mathrel{\mathtt{!=}}$ \textcolor{myred}{\small{\texttt{[Unrelevant]}}}}{Add $r_d$ to $w_d$\;} }}

\uElseIf{\ret~== \textcolor{myred}{\small{\texttt{[Entity Retrieval]}}}}{
Retrieve tail entities $\mathbf{E}$ using $r_d$\;
\mgen predicts \cre\;
\ForEach{$e \in \mathbf{E}$}{\mgen predicts \crel\;
 \uIf{\crel $\mathrel{\mathtt{!=}}$ \textcolor{myred}{\small{\texttt{[Unrelevant]}}}}{Add $e_d$ to $w_d$\;} }

 }
\uElseIf{\ret~== \textcolor{myred}{\small{\texttt{[No Retrieval]}}}}{
\mgen predicts \cuse given $q, w_{<d}$\;
} 
Rank $w_d$ based on reflection score and append $w_d$ to $T$\;
}
\end{algorithm}
For special token incorporation, we modularize each task with different instruction prompts and leverage critic model $\mathcal{C}$ like GPT models\footnote{\url{https://platform.openai.com/docs/models}} for assessment, facilitating the efficient insertion of reflection tokens. We prompt the critic model with type-specific instructions, wherein relevant knowledge from the reasoning path is extracted and provided as input. Through few-shot demonstrations, the model produces the corresponding evaluation and reflection. For each reasoning path, we divide it into segments based on the number of hops. At each hop, we assess the relevance of both candidate entities and relations, while simultaneously evaluating the logical coherence of the current path. Retrieval tokens are inserted accordingly until the answer entity is reached. The utility of the final answer is evaluated at last. Ultimately, each reasoning path is assembled sequentially with self-reflection tokens. Notably, when the model predicts the reasoning as \textit{[Unreasonable]}, the retrieval process is immediately terminated, and the answer is added directly. This ensures that the training data avoids propagation through unreasonable paths. Moreover, the model can learn to provide responses based on its own knowledge when encountering unreasonable paths. A compact description of the critic annotation protocol is provided in Appendix~\ref{app:prompt}. 
\paragraph{Generator Learning.}
We train the generator model \mgen by training on the curated reasoning paths augmented with self-reflection tokens from $\mathcal{D}_{gen}$. We approximate the expectation with $K$ sampled valid paths $\mathcal{W}_k^* \subseteq  \mathcal{W}^*$, using the standard next token objective:
\begin{equation}\label{eq:gen_training}
\begin{aligned}
\hspace{-2mm}
   &\mathcal{L} = \max_{\theta}\mathbb{E}_{(w,q, r)} \log \left( P_{\theta}\left( r | w, q\right)P_{\theta}\left(w|q, \mathcal{G}\right)\right) \\ 
 &\propto \sum_{w\in\mathcal{W}_k^*}\sum_{i=1}^D\log \left( P_{\theta}\left( r_i|w_{\leq i}\right) P_{\theta}\left(w_i|w_{<i},q,\mathcal{G} \right) \right),
    \end{aligned}
    \end{equation}
where $(w,q, r)$ is sampled from the distribution of $\mathcal{D}_{gen}$. The generator model \mgen learns to predict the next-hop path accompanied by the corresponding reflection tokens. During training, the loss is computed jointly for the retrieved knowledge included in the training sequence, enabling the model to concurrently learn the implicit mapping between queries and structural knowledge. The original vocabulary $\mathcal{V}$ is expanded with all reflection tokens.
\subsection{\model Inference}
Fine-grained reflection tokens provide a quantitative evaluation of the reasoning path. During the inference process, we employ a tree-based reasoning framework and assign a score derived from generation probability to each node. The scoring mechanism enables the effective pruning of redundant nodes. Additionally, by incorporating a hypothesis-enhancement method, we improve retrieval accuracy by actively generating candidates one future step forward during the relation retrieval process.
\paragraph{Hypo-Generator.}
\label{sec:hypo}
Direct retrieval based solely on coarse-grained retrievers or LLMs often suffers from low precision due to the inherent gap between the query and the underlying knowledge~\cite{query-rewrite, zhang2024alter}. Moreover, the representation of relations (hierarchical in Freebase) in graphs does not always align with natural language. The issue is even more severe in automatically-constructed knowledge graphs~\cite{codeKG,li-etal-2022-c3kg}. During the training phase, \model learns the implicit associations between queries and structural knowledge by actively predicting retrieved relations from the graph. In the relation retrieval process during inference, the trained model predicts one more step for hypothetical relations. The predictions are transformed as additional input to the retriever. The output from the hypo-generator is aligned with the representation of relations in the graph, effectively bridging the gap between query representations and structured knowledge.
\paragraph{Tree-based inference.}
During each retrieval step, the generator model is capable of processing multiple candidates (both relations and entities) in parallel, leading to the generation of diverse downstream reasoning paths. The parallel candidate sets collectively form a reasoning tree for the given query. Tree-based reasoning methods~\cite{yao2024tree, feng2023alphazero,wu2024comparative} have recently been widely adopted to enhance the reasoning process. Unlike other approaches that train a separate reward model, our model leverages the evaluation of special tokens as a process reward model, enabling effective assessment of tree nodes within the graph. During inference time, we integrate both hard and soft constraints and adopt a hop-level beam search strategy, retaining the top-$B$ candidates with the highest relevance and logical coherence scores. Specifically, for any candidate with undesirable tokens generated (\eg \textit{Unrelevant}), we simply prune it. Otherwise, we proceed to explore based on the current candidate and compute scores derived from the generation probability of special tokens. For the leaf nodes of the tree, we traverse back through the entire tree to aggregate final scores. The detailed score mechanism can be found in Appendix~\ref{app:score}.

\section{Experiments}
\subsection{Datasets and Evaluation Metrics.} We evaluate our proposed method and compare it with other methods on two widely-used KGQA benchmark datasets: WebQSP~\citep{STAGG} and CWQ~\citep{CWQ}. Specifically, WebQSP contains 4,737 natural language questions with SPARQL queries, with 3,098 in the training set and 1,639 in the testing set. CWQ contains 34,689 natural language questions with SPARQL queries, with 27,639 in the training set  3,519 in the validation set and 3,531 in the testing set. Both datasets are based on the Freebase~\citep{Freebase} KB.
\subsection{Baselines.} We compare \model with 13 baseline methods grouped into 3 categories: 1) Semantic Parsing (SP)-based methods, 2) Information Retrieval (IR)-based methods, and 3) LLM-based methods.

\subsection{Experimental Settings.}
\paragraph{Training details.}
In total, we collect 29,117 training samples based on the training splits of WebQSP and CWQ. For the modular evaluation task of different self-reflection tokens, we employ GPT-4o-mini considering computational costs and efficiency. During the training process, we use 4 NVIDIA A100 GPUs with 40GB memory to train the generator model. All models are trained for 3 epochs, with a batch size of 16 and a maximum learning rate of 1e-5. We use Deepspeed stage 3~\cite{rajbhandari2020zero} for multi-GPU distributed training. For the efficient training framework, we utilize LlamaFactory~\cite{zheng-etal-2024-llamafactory}. We employ LLama3-8b~\cite{dubey2024llama} as our base LLM backbone. 
\paragraph{Inference settings.}
For each hop-level segment, we employ a beam width of 3 by default. For the WebQSP dataset, the default search depth is set to 2, while for CWQ, the default search depth is 4. During the inference process, we utilize VLLM to accelerate reasoning. For the default retriever $\mathcal{R}$, we adopt bge-large-en-v1.5~\cite{bge_embedding}. During the construction of training data and the retrieval process, the default number of retrieved items is set to $K=5$.

\begin{table}[]
\caption{\small{Results of different methods on WebQSP and CWQ. (We use \underline{underline} to denote the second-best performance, \textbf{bold} to denote the best performance. $B$ stands for beam-width and \textit{Exhausted} means exhausted search.)}}
\centering
\footnotesize
\renewcommand{\arraystretch}{0.90}
\setlength{\tabcolsep}{2.5pt}
\setlength{\aboverulesep}{0.25ex}
\setlength{\belowrulesep}{0.25ex}
\begin{tabular}{@{}lcc@{}}
\toprule
\multicolumn{1}{c}{\multirow{2}{*}{Model}} & \multicolumn{2}{c}{Hit@1 (\%)} \\ \cmidrule(l){2-3} 
\multicolumn{1}{c}{} & \textbf{\textsc{WebQSP}} & \textbf{\textsc{CWQ   }} \\ \midrule
\rowcolor[rgb]{0.9,0.9,0.9}
\multicolumn{3}{c}{\textit{$\clubsuit$ SP-based methods}} \\ \midrule
DECAF~\cite{DECAF} & 82.1 & 70.4  \\
TIARA~\cite{shu-etal-2022-tiara} &  75.2 &  -  \\
ArcaneQA~\cite{gu-su-2022-arcaneqa} & 75.6&  -\\
ChatKBQA~\cite{luo-etal-2024-chatkbqa} &86.4 & \textbf{86.0}\\
\rowcolor[rgb]{0.9,0.9,0.9}
\multicolumn{3}{c}{\textit{$\heartsuit$ IR-based methods}} \\ \midrule
GrafNet~\cite{sun-etal-2018-open} & 66.4 &   36.8\\
PullNet~\cite{sun2019pullnet} &  68.1&   45.9\\
Subgraph Retrieval~\cite{zhang-etal-2022-subgraph} & 69.5 &  50.2 \\ 
RNG-KBQA~\cite{ye-etal-2022-rng} & 71.1 & - \\
UniKGQA~\cite{UniKGQA} & 77.2& 51.2\\
\midrule
\rowcolor[rgb]{0.9,0.9,0.9}
\multicolumn{3}{c}{\textit{$\spadesuit$ LLM-based methods}} \\ \midrule
 StructGPT~\cite{jiang-etal-2023-structgpt} &  72.6 &   -\\
 Interactive KBQA~\cite{xiong-etal-2024-interactive} & 71.2 & 59.2 \\
 KG-Agent~\cite{jiang2024kg} & 83.3 & 72.2\\
 RoG~\cite{luo2023reasoning} & 85.7 &  62.6\\
ToG w/ChatGPT~\cite{sun2023think} &  76.2&  57.1\\
ToG w/GPT-4~\cite{sun2023think} &  82.6&  67.6\\
\model ($B=1$)&  84.0 &   61.7 \\ 
\model ($B=3$)& \underline{90.2} &   72.4 \\ 
\model (\textit{Exhausted}) & \textbf{93.5} & \underline{79.8} \\
\bottomrule
\end{tabular}
\label{tab: main}
\end{table}

\subsection{Main Results}
We present in Table~\ref{tab: main} the comparative results of \model employing beam-width of $1$ and $3$, as well as exhaustive search, with the baselines on the WebQSP and CWQ datasets. From the table, it is evident that our method achieves significant improvements on both datasets. Notably, \model achieves state-of-the-art results on WebQSP. On CWQ, the model's performance surpasses all IR-based and LLM-based models. Specifically, when compared to ToG, the latest graph-based iterative reasoning approach leveraging GPT-4, \model ($B=3$) achieves improvements of $7.6\%$ and $4.8\%$, respectively. As the beam width increases, the performance of our model also improves. The outstanding performance of \model underscores its capability to effectively explore knowledge within the graph and accurately identify plausible answers.


\subsection{Ablation Study}
We carry out an ablation study to assess the impact of various components and hyperparameters on the performance of \model, as well as to explore the contribution of each type of self-reflection token.

\begin{figure}[htbp]
    \centering\includegraphics[width=\linewidth]{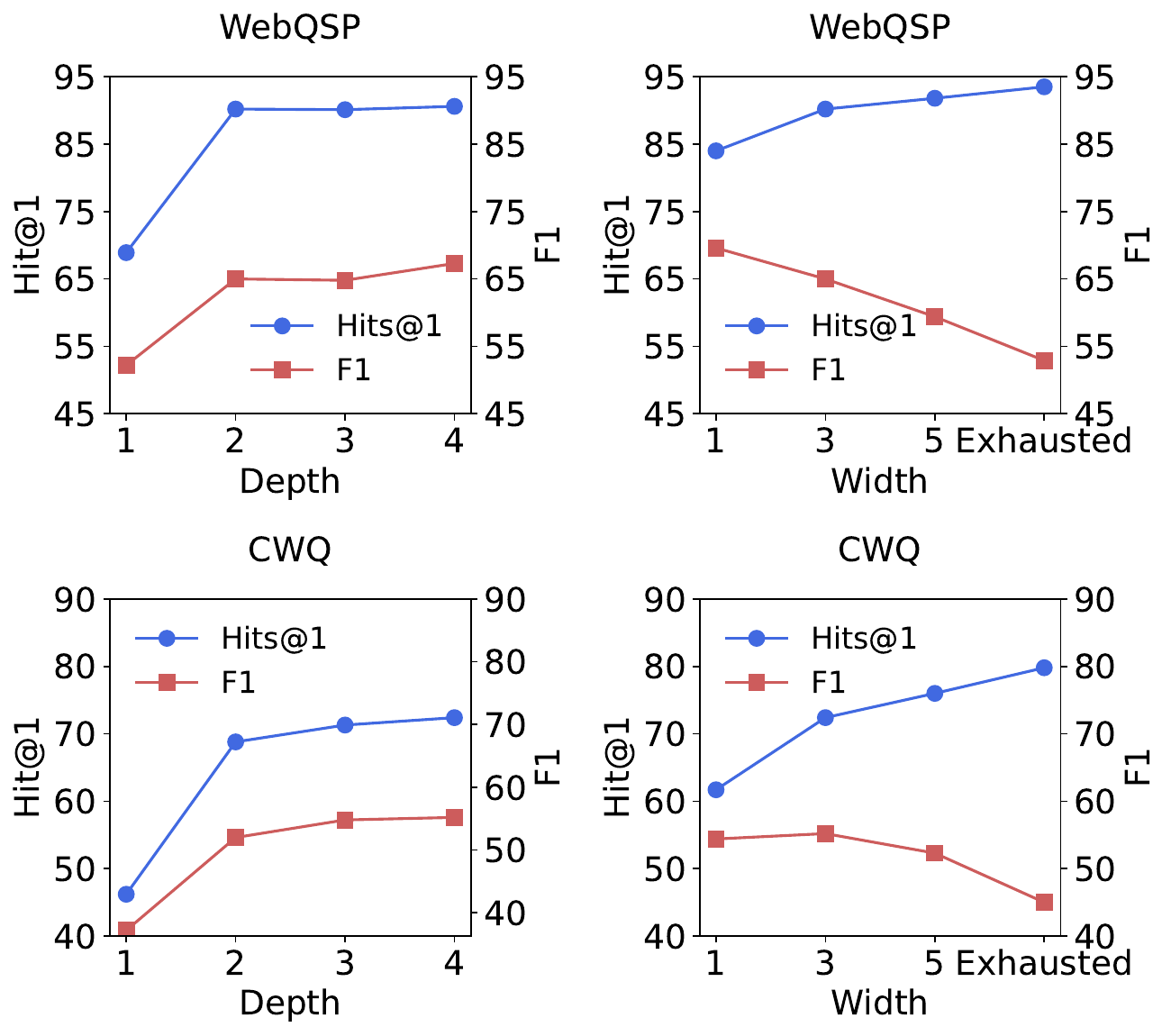}
    \caption{Ablation results of different reasoning depth and search depth on the WebQSP and CWQ.}
    \label{fig:abl-1}
\end{figure}
\paratitle{Analysis of Search Depth \& Width.}
To investigate the impact of reasoning depth and beam width on the performance of \model, we conduct ablation experiments on the CWQ and WebQSP datasets. We vary the depth from $1$ to $4$ and the beam-width from $1$ to $5$. Additionally, we adopt an exhaustive search during the retrieval process, retaining all valid nodes for expansion (excluding undesirable tokens). The results are presented in Figure~\ref{fig:abl-1}. From the results, it can be observed that increasing the width and depth during reasoning significantly enhances the performance, demonstrating that \model is capable of capturing deeper and broader effective information by improving the exploration range. For the WebQSP dataset, the decline in performance improvement when the hops exceed $2$ is attributed to the dataset's focus on questions primarily within 2 hops. Meanwhile, broader reasoning, particularly with a wider retrieval width, may introduce noise, leading to increased uncertainty, as reflected in the changes in F1 scores. Therefore, selecting appropriate parameters is indeed necessary in practice.

\paratitle{Analysis of Self-Reflection Tokens.}
We further investigate the role of self-reflection tokens in \model by conducting an ablation study on the WebQSP dataset. Specifically, during the reasoning process, we selectively removed the contributions of four types of scores, including \textit{Relevance}, \textit{Rationality}, \textit{Utility}, and \textit{Sequence}. The experimental results are illustrated in Table~\ref{abl:special}. From the experimental results, we observe that removing either the \textit{Relevance} or \textit{Rationality} score individually leads to a slight decline in model performance. However, when both \textit{Relevance} and \textit{Rationality} scores are omitted, the model's performance deteriorates significantly. This demonstrates the importance of both \textit{Relevance} and \textit{Rationality} and the complementary relationship of the two aspects. The most significant performance decline occurs when only the utility token is utilized. In this scenario, the model solely evaluates the quality of generations, which compromises the factuality of the reasoning process. With all scores integrated together, \model achieves optimal performance.
\begin{table}[]
\caption{Ablation results for investigating the impact of special tokens, the check mark indicates the retention of the contribution of that particular token type to the final score.}
\resizebox{1\linewidth}{!}{
\renewcommand{\arraystretch}{1}
\setlength{\tabcolsep}{3pt}

\begin{tabular}{@{}cccc|c@{}}
\toprule
Relevance & Rationality & Utility & Sequence & \textbf{\textsc{WebQSP}} \\ \midrule
\usym{1F5F8} & \usym{1F5F8} & \usym{1F5F8}& \usym{1F5F8} &  83.9\\
\textbf{} &\usym{1F5F8} & \usym{1F5F8} &\usym{1F5F8} & 82.4 \scriptsize{\textcolor{cadmiumgreen}{(-1.5)}}\\ 
\usym{1F5F8} & \textbf{} & \usym{1F5F8} & \usym{1F5F8} & 83.6\scriptsize{\textcolor{cadmiumgreen}{(-0.3)}} \\
\textbf{} & \textbf{} & \usym{1F5F8} & \usym{1F5F8} & 80.9\scriptsize{\textcolor{cadmiumgreen}{(-3.0)}} \\
\textbf{} & \textbf{} & \usym{1F5F8} & \textbf{} & 78.9\scriptsize{\textcolor{cadmiumgreen}{(-5.0)}} \\
\textbf{} & \textbf{} & \textbf{} & \usym{1F5F8} & 79.7\scriptsize{\textcolor{cadmiumgreen}{(-4.2)}} \\
\bottomrule
\end{tabular}
}
\label{abl:special}
\end{table}

\begin{table}[]
\small
\vskip -0.05in
\caption{Ablation results of different retrieval settings, where \wo
hypo-generator represents using the query directly as the retrieval input.}
\centering
\resizebox{0.9\linewidth}{!}{
\renewcommand{\arraystretch}{1}
\setlength{\tabcolsep}{3pt}
\begin{tabular}{@{}lcc@{}}
\toprule
\multicolumn{1}{c}{\multirow{2}{*}{Methods}} & \multicolumn{2}{c}{Hit@1} \\ \cmidrule(l){2-3} 
 & \textbf{\textsc{WebQSP}} & \textbf{\textsc{CWQ}} \\ \midrule
Ours ($K=5$) & 90.2 & 72.4 \\ \midrule
\wo hypo generator &  85.1 &  69.3\\
\wo naive retriever & 87.2 & 70.3 \\
\model ($K=3$) &  89.5 &  71.1\\ \bottomrule
\end{tabular}
}
\label{abl:hypo}
\vskip -0.1in
\end{table}
\paratitle{Analysis of Hypo-Generator.}
We conducted an investigation into the impact of the hypo-generator module on the retrieval efficiency of \model. The experimental results are presented in the Table~\ref{abl:hypo}. We compared the performance of direct retrieval based solely on the query and retrieval enhanced exclusively by the hypo-generator. From the results, it can be observed that both approaches exhibit some performance degradation. However, it is noteworthy that \model with $K=3$ significantly outperforms the individual use of either method. This indicates that the hypo-generator is effectively integrated into the \model framework and serves to enhance the retrieval process.

\subsection{Additional Analysis}
\paratitle{Retrieval Evaluation.}
We further evaluate relation-retrieval recall on WebQSP against the default retriever and ToG. \model consistently achieves higher recall across retrieval depths; at top-$k=20$, it reaches 87.9\% recall, compared with 79.0\% for ToG and 66.9\% for the default retriever. This directly supports the effectiveness of the learned retrieval mechanism.

\paratitle{Efficiency Analysis.}
We further compare inference latency with ToG under the same experimental conditions on 300 randomly sampled WebQSP examples. ToG, with a search width of 5 and depth of 2, requires 71.82 seconds on average. In contrast, \model with $B=5$ and $D=2$ takes 12.7 seconds, while exhaustive search takes 27.1 seconds. Thus, even exhaustive search is substantially faster than ToG while maintaining better performance.

\paratitle{Human Evaluation of the Critic Model.}
We manually assess 30 randomly sampled WebQSP instances and compare critic-model outputs with human judgments. For relation relevance, entity relevance, and rationality, GPT-4o-mini achieves normalized agreement scores of 0.82, 0.93, and 0.91, respectively, compared with 0.81, 0.95, and 0.88 for GPT-3.5. The close agreement with human evaluation supports the quality of the automatically collected reflection supervision; we use GPT-4o-mini for its lower cost and higher throughput.

\subsection{Transferring to Other KGs}
To further validate the transferability of our approach to other KGs, we conduct additional experiments on the Wiki-Movies KG, whose relational representations differ from those of Freebase. We use the MetaQA-3hop~\cite{zhang2018variational} dataset for the construction of training data and evaluation purposes. From the training split, we sample 1,000 examples and construct the training set following the same data construction methodology in Section~\ref{sec:train}. Subsequently, we fine-tune \model using two distinct training strategies, as utilized by RoG~\cite{luo2023reasoning}: 1) training from scratch, where the model is trained directly from the base LLM; and 2) transfer from Freebase, where the model pre-trained on Freebase is further fine-tuned on the new dataset. We select several representative works that demonstrate transferability across multiple knowledge graphs for comparison. The comparative results are presented in Table~\ref{exp:transfer}. From the results, it can be observed that our model demonstrates adaptability to different KGs. Moreover, the model fine-tuned from Freebase exhibits superior performance, demonstrating that by learning structural knowledge representation internally, the model can rapidly transfer to other new graphs.
\subsection{Case Study}
\begin{table}[]
\caption{Results of transferability of \model on MetaQA-3hop dataset based on Wiki-Movies KG. }
\resizebox{1\linewidth}{!}{
\renewcommand\arraystretch{0.9}
\setlength\tabcolsep{4pt}
\centering
\begin{tabular}{@{}lc@{}}
\toprule
\multicolumn{1}{c}{\multirow{2}{*}{Methods}} & \multicolumn{1}{c}{Hit@1 (\%)} \\ \cmidrule(l){2-2} 
\multicolumn{1}{c}{} & \multicolumn{1}{c}{\textbf{\textsc{MetaQA-3hop}}} \\ \midrule
StructGPT~\cite{jiang-etal-2023-structgpt} &  80.2\\
Retrieval \& Reasoning~\cite{ji-etal-2024-retrieval} & 76 \\
RoG (train from scratch)~\cite{luo2023reasoning} &  84.81\\
RoG (transfer from Freebase)~\cite{luo2023reasoning} &  88.98\\
\midrule
Ours (train from scratch)& 87.7 \\ 
Ours (transfer from Freebase)& \textbf{89.4}   \\ \bottomrule
\end{tabular}
}
\vskip -0.1in
\label{exp:transfer}
\end{table}
Three WebQSP cases compare \model with ToG; detailed paths and reflection labels appear in Appendix~\ref{app:cases}.
For Niall Ferguson's spouse, both methods identify Ayaan Hirsi Ali, but \model terminates after two hops and prunes an unreasonable Marriage branch, whereas ToG continues to depth 3.

For Marc Chagall's art, \model retrieves five art movements through the associated-movement relation, whereas ToG returns the generic answer ``Painting''.

For Florida Panthers, \model returns the venue entities Miami Arena and BB\&T Center, while ToG answers ``Sunrise'' despite retrieving Miami Arena. These cases illustrate early stopping and targeted relation/entity selection.

\section{Conclusion}
This work introduces \model, a novel framework designed for reliable reasoning over structured graph data through on-demand retrieval and self-reflection mechanisms. By leveraging reasoning paths as weakly supervised training data, the model is trained to perform iterative retrieval, reflection, and generation. Furthermore, the retrieval performance is significantly enhanced through the hypo-generator, enabling the capture of latent information within the graph. The generated reasoning paths exhibit high interpretability. We posit that our approach offers valuable insights into how models comprehend structured information, contributing to the broader exploration of interpretable reasoning in machine learning.
\section{Limitations}
\model achieves iterative reflective reasoning over structured graphs in an end-to-end manner, where the final answer is derived through a reasoning tree during the inference process. Recently, tree-based reasoning methods are frequently employed in complex reasoning tasks; however, conventional tree-based approaches often introduce uncertainty, which is reflected in suboptimal precision. In the future, integrating advanced tree reasoning techniques, such as those utilized in Reinforcement Learning from Human Feedback (RLHF) and deep reasoning models~\cite{feng2023alphazero, xie2023self,guan2025rstar}, could further enhance the precision of tree-structured reasoning. We will leave these explorations for future work.

\appendix
\section{Additional Method Details}
\label{sec:appendix}

\subsection{Reflection Token Definitions}
\label{app:tokens}
\begin{sloppypar}
The generator uses four types of reflection tokens. The \emph{retrieval} token decides whether to stop retrieval or retrieve relations/entities: \texttt{[No Retrieval]}, \texttt{[Relation Retrieval]}, and \texttt{[Entity Retrieval]}. The \emph{relevance} token evaluates retrieved relations or entities as \texttt{[Fully Relevant]}, \texttt{[Partially Relevant]}, or \texttt{[Unrelevant]}. The \emph{rationality} token evaluates the current reasoning path as \texttt{[Fully Reasonable]}, \texttt{[Partially Reasonable]}, or \texttt{[Unreasonable]}. The \emph{utility} token scores the final answer from \texttt{[Utility:1]} to \texttt{[Utility:5]}.
\end{sloppypar}

\subsection{Training Data Construction}
\label{app:data-creation}
Algorithm~\ref{algo:data_gen} summarizes how valid KG reasoning paths are augmented with retrieval and reflection supervision. At each hop, the retriever forms relation and entity candidates, while the critic assigns relevance and rationality labels. An \texttt{[Unreasonable]} prediction terminates expansion of the current path. After reaching an answer, the critic assigns a utility token.

\begin{algorithm}[t]
\small
\caption{$\mathcal{M}_{gen}$ Data Creation}\label{algo:data_gen}
\KwIn{Query $q$; valid path $w=w_{1:D}$; retriever $\mathcal{R}$; critic $\mathcal{C}$}
\KwOut{Reasoning path augmented with reflection tokens}
\For{$d\leftarrow1$ \KwTo $D$}{
  $(r_d,e_d)\leftarrow w_d$; add \rret\;
  $\mathbf{C_r}\leftarrow\mathrm{Retrieve}(q,r_d,e_d;\mathcal{R})$; $\mathcal{C}$ predicts \crel\ for $\mathbf{C_r}$\;
  add \eret; $\mathbf{C_e}\leftarrow\mathrm{SearchTailNode}(e_d,r_d)$; $\mathcal{C}$ predicts \crel\ for $\mathbf{C_e}$\;
  $\mathcal{C}$ predicts \cre\ from $w_{\le d}$\;
  \If{\cre\ is \texttt{[Unreasonable]}}{\textbf{break}\;}
}
add \texttt{[No}\allowbreak\texttt{ Retrieval]}; $\mathcal{C}$ predicts \cuse\ for the final answer\;
\end{algorithm}

\subsection{Inference Score}
\label{app:score}
For a reflection token $\hat t$ generated at depth $d$, we normalize the logit $p_d(\hat t)$ within the $G$ categories of the same token type:
\begin{equation}
 s_d(\hat t)=\frac{\exp(p_d(\hat t))}{\sum_{i=1}^{G}\exp(p_d(t_i))}.
\end{equation}
The score of a node is
\begin{equation}
 S(d)=S_{\mathrm{seq}}(d)+k_1S_{\mathrm{rel}}(d)S_{\mathrm{rat}}(d)+k_2S_{\mathrm{uti}},
\end{equation}
where $k_1=1$ and $k_2=0.5$. We use
\begin{align}
S_{\mathrm{rel}}(d) &= s_d(\texttt{Fully Relevant})\notag\\
&\quad + 0.5s_d(\texttt{Partially Relevant}),\\
S_{\mathrm{rat}}(d) &= s_d(\texttt{Fully Reasonable})\notag\\
&\quad + 0.5s_d(\texttt{Partially Reasonable}).
\end{align}
and compute $S_{\mathrm{uti}}$ from the probability distribution over the five utility levels at the final depth. Leaf-path scores are obtained by backtracking and aggregating node scores along the path.

\subsection{Critic Prompting}
\label{app:prompt}
We present the instructions used for collecting self-reflection tokens, including \textit{Relevance}, \textit{Rationality}, and \textit{Utility}. 
\begin{tcolorbox}[breakable,colback=white,colframe=black!40,boxrule=0.4pt,arc=0pt,boxsep=0pt,left=1mm,right=1mm,top=1mm,bottom=1mm,before skip=2pt,after skip=2pt]
\footnotesize
\lstset{basicstyle=\rmfamily\footnotesize,columns=fullflexible,aboveskip=0pt,belowskip=2pt}
\textbf{Relation Relevance}\par\nobreak
\begin{lstlisting}
You will receive a query, topic entity, evidence and optional preceding sentences containing history information. The evidence contains graph relationships possibly useful to answering the query. Your task is evaluate each relationship's contribution to answering the query and provide a relevance score for each relation, output your explanations for the score.
The score of relevance range from [Fully Relevant], [Partially Relevant] to [Unrelevant]:
- If the relationship directly contains information directly about the query or can answer the query with information in preceding sentences, return [Fully Relevant].
- If the relationship do not directly answer the query, but includes information possibly point to the answer, return [Partially Relevant].
- If the relationship contains irrelevant information about the query, return [Unrelevant].
\end{lstlisting}

\noindent{\color{black!30}\rule{\linewidth}{0.3pt}}\par\nobreak
\textbf{Entity Relevance}\par\nobreak
\begin{lstlisting}
You will receive a query, evidence and optional preceding historical information for the task. The evidence and preceding information include associated retrieved knowledge graph triplets presented as (head entity, relation, tail entity).
Your task is to assign a relevance score to the query for each tail entity in the evidence. Additionally, you are required to provide explanations for the scores assigned.
The relevance scores should fall into one of the following categories: [Fully Relevant], [Partially Relevant], or [Unrelevant].
\end{lstlisting}

\noindent{\color{black!30}\rule{\linewidth}{0.3pt}}\par\nobreak
\textbf{Path Rationality}\par\nobreak
\begin{lstlisting}
You will receive a query, output and a reasoning path. The reasoning path contains the current reasoning process starting from the topic entitiy to the answer. Your task is to rate rationality score for the path and output your explanations for the score.
The score of rationality range from [Fully Reasonable], [Partially Reasonable] to [Unreasonable].
\end{lstlisting}

\noindent{\color{black!30}\rule{\linewidth}{0.3pt}}\par\nobreak
\textbf{Answer Utility}\par\nobreak
\begin{lstlisting}
You will be given a query and the answers, where the answers may consist of one or more individual answers, separated by commas(,).
Your task is to generate a **rating** to evaluate whether the answer is a useful response to the query.
Use the following entailment scale to give the utility score:
[Utility:5]: Generally, the output provides a complete, highly detailed, and informative response to the query, fully satisfying the information needs.
[Utility:4]: Generally, the output mostly fulfills the need in the query and provides helpful answers, while there can be some minor improvements, such as discussing more detailed information or providing additional correct answers beyond the current output.
[Utility:3]: Generally, the output is correct and acceptable, but there are obvious problems, such as being too vague or not specific enough, limiting its helpfulness in addressing the query.
[Utility:2]: Generally, the output still discusses the topic of the query, but it is incorrect or does not actually meet the requirements of the query.
[Utility:1]: Generally, the output is completely irrelevant to the query or does not give an answer in the end.
\end{lstlisting}
\end{tcolorbox}

\section{Case Studies}
\label{app:cases}
These WebQSP examples retain the recorded answers and labels. Selected branches are shown, with relation names shortened to their final components.

\paragraph{Case 1: Early Termination.}
For ``Who is Niall Ferguson's wife?'', ToG reaches Ayaan Hirsi Ali via Thomas Ferguson at depth 2 but continues to depth 3. \model follows Niall Ferguson $\to$ \texttt{spouse\_s} $\to$ \texttt{m.0j4jq57} $\to$ Ayaan Hirsi Ali, moving from \texttt{[Partially Reasonable]} to \texttt{[Fully Reasonable]}. It stops Marriage $\to$ \texttt{type\_of\_union} $\to$ \texttt{m.0j4jt43} as \texttt{[Unreasonable]}. Both methods answer Ayaan Hirsi Ali; \model assigns \texttt{[Utility:5]}.

\paragraph{Case 2: Relation Selection.}
For ``What type of art does Marc Chagall do?'', ToG answers ``Painting''. Via \texttt{associated\_\allowbreak periods\_\allowbreak or\_\allowbreak movements}, \model retrieves Modern art as \texttt{[Fully Relevant]} and Cubism, Surrealism, Expressionism, and Fauvism as \texttt{[Partially Relevant]}. It answers with all five movements and receives \texttt{[Utility:3]}, illustrating a different relation choice rather than maximum utility.

\paragraph{Case 3: Entity Selection.}
For ``Where do Florida Panthers play?'', ToG retrieves Miami Arena at depth 2 but answers ``Sunrise''. Via \texttt{arena\_stadium}, \model rates Miami Arena \texttt{[Partially Relevant]} and BB\&T Center \texttt{[Fully Relevant]}, answering ``Miami Arena; BB\&T Center'' with \texttt{[Utility:4]}. This contrasts a location answer with the benchmark's venue entities.

\section*{GenAI Usage Disclosure}
GPT was used only for language polishing, including checking grammatical fluency and making minor wording edits to the manuscript. No generative AI tools were used in the research design, experiments, data analysis, or interpretation of results. All AI-assisted edits were reviewed and verified by the authors.

\bibliographystyle{ACM-Reference-Format}
\bibliography{custom}

@inproceedings{GrailQA,
author = {Gu, Yu and Kase, Sue and Vanni, Michelle and Sadler, Brian and Liang, Percy and Yan, Xifeng and Su, Yu},
title = {Beyond I.I.D.: Three Levels of Generalization for Question Answering on Knowledge Bases},
year = {2021},
isbn = {9781450383127},
publisher = {Association for Computing Machinery},
address = {New York, NY, USA},
url = {https://doi.org/10.1145/3442381.3449992},
doi = {10.1145/3442381.3449992},
booktitle = {Proceedings of the Web Conference 2021},
pages = {3477–3488},
numpages = {12},
location = {Ljubljana, Slovenia},
series = {WWW '21}
}

@InProceedings{SPARQL,
author="P{\'e}rez, Jorge
and Arenas, Marcelo
and Gutierrez, Claudio",
editor="Cruz, Isabel
and Decker, Stefan
and Allemang, Dean
and Preist, Chris
and Schwabe, Daniel
and Mika, Peter
and Uschold, Mike
and Aroyo, Lora M.",
title="Semantics and Complexity of SPARQL",
booktitle="The Semantic Web - ISWC 2006",
year="2006",
publisher="Springer Berlin Heidelberg",
address="Berlin, Heidelberg",
pages="30--43",
isbn="978-3-540-49055-5"
}

@inproceedings{ji-etal-2024-retrieval,
    title = "Retrieval and Reasoning on {KG}s: Integrate Knowledge Graphs into Large Language Models for Complex Question Answering",
    author = "Ji, Yixin  and
      Wu, Kaixin  and
      Li, Juntao  and
      Chen, Wei  and
      Zhong, Mingjie  and
      Jia, Xu  and
      Zhang, Min",
    editor = "Al-Onaizan, Yaser  and
      Bansal, Mohit  and
      Chen, Yun-Nung",
    booktitle = "Findings of the Association for Computational Linguistics: EMNLP 2024",
    month = nov,
    year = "2024",
    address = "Miami, Florida, USA",
    publisher = "Association for Computational Linguistics",
    url = "https://aclanthology.org/2024.findings-emnlp.446/",
    doi = "10.18653/v1/2024.findings-emnlp.446",
    pages = "7598--7610"
}

@article{sun2023think,
  title={Think-on-graph: Deep and responsible reasoning of large language model with knowledge graph},
  author={Sun, Jiashuo and Xu, Chengjin and Tang, Lumingyuan and Wang, Saizhuo and Lin, Chen and Gong, Yeyun and Shum, Heung-Yeung and Guo, Jian},
  journal={arXiv preprint arXiv:2307.07697},
  year={2023}
}

@inproceedings{jiang-etal-2023-structgpt,
    title = "{S}truct{GPT}: A General Framework for Large Language Model to Reason over Structured Data",
    author = "Jiang, Jinhao  and
      Zhou, Kun  and
      Dong, Zican  and
      Ye, Keming  and
      Zhao, Xin  and
      Wen, Ji-Rong",
    editor = "Bouamor, Houda  and
      Pino, Juan  and
      Bali, Kalika",
    booktitle = "Proceedings of the 2023 Conference on Empirical Methods in Natural Language Processing",
    month = dec,
    year = "2023",
    address = "Singapore",
    publisher = "Association for Computational Linguistics",
    url = "https://aclanthology.org/2023.emnlp-main.574",
    doi = "10.18653/v1/2023.emnlp-main.574",
    pages = "9237--9251",
}

@inproceedings{gu-etal-2023-dont,
    title = "Don`t Generate, Discriminate: A Proposal for Grounding Language Models to Real-World Environments",
    author = "Gu, Yu  and
      Deng, Xiang  and
      Su, Yu",
    editor = "Rogers, Anna  and
      Boyd-Graber, Jordan  and
      Okazaki, Naoaki",
    booktitle = "Proceedings of the 61st Annual Meeting of the Association for Computational Linguistics (Volume 1: Long Papers)",
    month = jul,
    year = "2023",
    address = "Toronto, Canada",
    publisher = "Association for Computational Linguistics",
    url = "https://aclanthology.org/2023.acl-long.270/",
    doi = "10.18653/v1/2023.acl-long.270",
    pages = "4928--4949",
}

@article{jiang2024kg,
  title={Kg-agent: An efficient autonomous agent framework for complex reasoning over knowledge graph},
  author={Jiang, Jinhao and Zhou, Kun and Zhao, Wayne Xin and Song, Yang and Zhu, Chen and Zhu, Hengshu and Wen, Ji-Rong},
  journal={arXiv preprint arXiv:2402.11163},
  year={2024}
}

@inproceedings{kb-binder,
    title = "Few-shot In-context Learning on Knowledge Base Question Answering",
    author = "Li, Tianle  and
      Ma, Xueguang  and
      Zhuang, Alex  and
      Gu, Yu  and
      Su, Yu  and
      Chen, Wenhu",
    editor = "Rogers, Anna  and
      Boyd-Graber, Jordan  and
      Okazaki, Naoaki",
    booktitle = "Proceedings of the 61st Annual Meeting of the Association for Computational Linguistics (Volume 1: Long Papers)",
    month = jul,
    year = "2023",
    address = "Toronto, Canada",
    publisher = "Association for Computational Linguistics",
    url = "https://aclanthology.org/2023.acl-long.385/",
    doi = "10.18653/v1/2023.acl-long.385",
    pages = "6966--6980",
}

@inproceedings{luo-etal-2024-chatkbqa,
    title = "{C}hat{KBQA}: A Generate-then-Retrieve Framework for Knowledge Base Question Answering with Fine-tuned Large Language Models",
    author = "Luo, Haoran  and
      E, Haihong  and
      Tang, Zichen  and
      Peng, Shiyao  and
      Guo, Yikai  and
      Zhang, Wentai  and
      Ma, Chenghao  and
      Dong, Guanting  and
      Song, Meina  and
      Lin, Wei  and
      Zhu, Yifan  and
      Luu, Anh Tuan",
    editor = "Ku, Lun-Wei  and
      Martins, Andre  and
      Srikumar, Vivek",
    booktitle = "Findings of the Association for Computational Linguistics: ACL 2024",
    month = aug,
    year = "2024",
    address = "Bangkok, Thailand",
    publisher = "Association for Computational Linguistics",
    url = "https://aclanthology.org/2024.findings-acl.122/",
    doi = "10.18653/v1/2024.findings-acl.122",
    pages = "2039--2056",
}

@inproceedings{jin-etal-2024-graph,
    title = "Graph Chain-of-Thought: Augmenting Large Language Models by Reasoning on Graphs",
    author = "Jin, Bowen  and
      Xie, Chulin  and
      Zhang, Jiawei  and
      Roy, Kashob Kumar  and
      Zhang, Yu  and
      Li, Zheng  and
      Li, Ruirui  and
      Tang, Xianfeng  and
      Wang, Suhang  and
      Meng, Yu  and
      Han, Jiawei",
    editor = "Ku, Lun-Wei  and
      Martins, Andre  and
      Srikumar, Vivek",
    booktitle = "Findings of the Association for Computational Linguistics: ACL 2024",
    month = aug,
    year = "2024",
    address = "Bangkok, Thailand",
    publisher = "Association for Computational Linguistics",
    url = "https://aclanthology.org/2024.findings-acl.11/",
    doi = "10.18653/v1/2024.findings-acl.11",
    pages = "163--184",
}

@inproceedings{xiong-etal-2024-interactive,
    title = "Interactive-{KBQA}: Multi-Turn Interactions for Knowledge Base Question Answering with Large Language Models",
    author = "Xiong, Guanming  and
      Bao, Junwei  and
      Zhao, Wen",
    editor = "Ku, Lun-Wei  and
      Martins, Andre  and
      Srikumar, Vivek",
    booktitle = "Proceedings of the 62nd Annual Meeting of the Association for Computational Linguistics (Volume 1: Long Papers)",
    month = aug,
    year = "2024",
    address = "Bangkok, Thailand",
    publisher = "Association for Computational Linguistics",
    url = "https://aclanthology.org/2024.acl-long.569/",
    doi = "10.18653/v1/2024.acl-long.569",
    pages = "10561--10582",
}

@inproceedings{zhang-etal-2022-subgraph,
    title = "Subgraph Retrieval Enhanced Model for Multi-hop Knowledge Base Question Answering",
    author = "Zhang, Jing  and
      Zhang, Xiaokang  and
      Yu, Jifan  and
      Tang, Jian  and
      Tang, Jie  and
      Li, Cuiping  and
      Chen, Hong",
    editor = "Muresan, Smaranda  and
      Nakov, Preslav  and
      Villavicencio, Aline",
    booktitle = "Proceedings of the 60th Annual Meeting of the Association for Computational Linguistics (Volume 1: Long Papers)",
    month = may,
    year = "2022",
    address = "Dublin, Ireland",
    publisher = "Association for Computational Linguistics",
    url = "https://aclanthology.org/2022.acl-long.396/",
    doi = "10.18653/v1/2022.acl-long.396",
    pages = "5773--5784",
}

@article{BM25,
author = {Robertson, Stephen and Zaragoza, Hugo},
title = {The Probabilistic Relevance Framework: BM25 and Beyond},
year = {2009},
issue_date = {April 2009},
publisher = {Now Publishers Inc.},
address = {Hanover, MA, USA},
volume = {3},
number = {4},
issn = {1554-0669},
url = {https://doi.org/10.1561/1500000019},
doi = {10.1561/1500000019},
journal = {Found. Trends Inf. Retr.},
month = {apr},
pages = {333–389},
numpages = {57}
}

@article{Contriever,
title={Unsupervised Dense Information Retrieval with Contrastive Learning},
author={Gautier Izacard and Mathilde Caron and Lucas Hosseini and Sebastian Riedel and Piotr Bojanowski and Armand Joulin and Edouard Grave},
journal={Transactions on Machine Learning Research},
issn={2835-8856},
year={2022},
url={https://openreview.net/forum?id=jKN1pXi7b0},
note={}
}

@article{luo2023reasoning,
  title={Reasoning on graphs: Faithful and interpretable large language model reasoning},
  author={Luo, Linhao and Li, Yuan-Fang and Haffari, Gholamreza and Pan, Shirui},
  journal={arXiv preprint arXiv:2310.01061},
  year={2023}
}

@inproceedings{sun-etal-2018-open,
    title = "Open Domain Question Answering Using Early Fusion of Knowledge Bases and Text",
    author = "Sun, Haitian  and
      Dhingra, Bhuwan  and
      Zaheer, Manzil  and
      Mazaitis, Kathryn  and
      Salakhutdinov, Ruslan  and
      Cohen, William",
    editor = "Riloff, Ellen  and
      Chiang, David  and
      Hockenmaier, Julia  and
      Tsujii, Jun{'}ichi",
    booktitle = "Proceedings of the 2018 Conference on Empirical Methods in Natural Language Processing",
    month = oct # "-" # nov,
    year = "2018",
    address = "Brussels, Belgium",
    publisher = "Association for Computational Linguistics",
    url = "https://aclanthology.org/D18-1455/",
    doi = "10.18653/v1/D18-1455",
    pages = "4231--4242",
}

@inproceedings{shu-etal-2022-tiara,
    title = "{TIARA}: Multi-grained Retrieval for Robust Question Answering over Large Knowledge Base",
    author = {Shu, Yiheng  and
      Yu, Zhiwei  and
      Li, Yuhan  and
      Karlsson, B{\"o}rje  and
      Ma, Tingting  and
      Qu, Yuzhong  and
      Lin, Chin-Yew},
    editor = "Goldberg, Yoav  and
      Kozareva, Zornitsa  and
      Zhang, Yue",
    booktitle = "Proceedings of the 2022 Conference on Empirical Methods in Natural Language Processing",
    month = dec,
    year = "2022",
    address = "Abu Dhabi, United Arab Emirates",
    publisher = "Association for Computational Linguistics",
    url = "https://aclanthology.org/2022.emnlp-main.555/",
    doi = "10.18653/v1/2022.emnlp-main.555",
    pages = "8108--8121",
}

@inproceedings{miller-etal-2016-key,
    title = "Key-Value Memory Networks for Directly Reading Documents",
    author = "Miller, Alexander  and
      Fisch, Adam  and
      Dodge, Jesse  and
      Karimi, Amir-Hossein  and
      Bordes, Antoine  and
      Weston, Jason",
    editor = "Su, Jian  and
      Duh, Kevin  and
      Carreras, Xavier",
    booktitle = "Proceedings of the 2016 Conference on Empirical Methods in Natural Language Processing",
    month = nov,
    year = "2016",
    address = "Austin, Texas",
    publisher = "Association for Computational Linguistics",
    url = "https://aclanthology.org/D16-1147/",
    doi = "10.18653/v1/D16-1147",
    pages = "1400--1409"
}

@inproceedings{das-etal-2017-question,
    title = "Question Answering on Knowledge Bases and Text using Universal Schema and Memory Networks",
    author = "Das, Rajarshi  and
      Zaheer, Manzil  and
      Reddy, Siva  and
      McCallum, Andrew",
    editor = "Barzilay, Regina  and
      Kan, Min-Yen",
    booktitle = "Proceedings of the 55th Annual Meeting of the Association for Computational Linguistics (Volume 2: Short Papers)",
    month = jul,
    year = "2017",
    address = "Vancouver, Canada",
    publisher = "Association for Computational Linguistics",
    url = "https://aclanthology.org/P17-2057/",
    doi = "10.18653/v1/P17-2057",
    pages = "358--365",
}

@inproceedings{guu2020retrieval,
  title={Retrieval augmented language model pre-training},
  author={Guu, Kelvin and Lee, Kenton and Tung, Zora and Pasupat, Panupong and Chang, Mingwei},
  booktitle={International Conference on Machine Learning},
  year={2020},
  url={https://dl.acm.org/doi/pdf/10.5555/3524938.3525306},
}

@inproceedings{jiang-etal-2023-active,
    title = "Active Retrieval Augmented Generation",
    author = "Jiang, Zhengbao  and
      Xu, Frank  and
      Gao, Luyu  and
      Sun, Zhiqing  and
      Liu, Qian  and
      Dwivedi-Yu, Jane  and
      Yang, Yiming  and
      Callan, Jamie  and
      Neubig, Graham",
    editor = "Bouamor, Houda  and
      Pino, Juan  and
      Bali, Kalika",
    booktitle = "Proceedings of the 2023 Conference on Empirical Methods in Natural Language Processing",
    month = dec,
    year = "2023",
    address = "Singapore",
    publisher = "Association for Computational Linguistics",
    url = "https://aclanthology.org/2023.emnlp-main.495/",
    doi = "10.18653/v1/2023.emnlp-main.495",
    pages = "7969--7992"
}

@article{asai2023selfrag,
      author    = {Asai, Akari and Wu, Zeqiu and Wang, Yizhong and Sil, Avirup and Hajishirzi, Hannaneh},
      title     = {{Self-RAG}: Learning to Retrieve, Generate, and Critique through Self-Reflection},
      year      = {2023},
     journal    = {arXiv preprint arXiv:2310.11511},
     url        = {https://arxiv.org/abs/2310.11511}
    }

@article{mavromatis2024gnn,
  title={GNN-RAG: Graph Neural Retrieval for Large Language Model Reasoning},
  author={Mavromatis, Costas and Karypis, George},
  journal={arXiv preprint arXiv:2405.20139},
  year={2024}
}

@inproceedings{wang2021relational,
  title     = {Relational message passing for knowledge graph completion},
  author    = {Wang, Hongwei and Ren, Hongyu and Leskovec, Jure},
  booktitle = {Proceedings of the 27th ACM SIGKDD Conference on Knowledge Discovery \& Data Mining},
  pages     = {1697--1707},
  year      = {2021}
}

@inproceedings{query-rewrite,
    title = "Query Rewriting in Retrieval-Augmented Large Language Models",
    author = "Ma, Xinbei  and
      Gong, Yeyun  and
      He, Pengcheng  and
      Zhao, Hai  and
      Duan, Nan",
    editor = "Bouamor, Houda  and
      Pino, Juan  and
      Bali, Kalika",
    booktitle = "Proceedings of the 2023 Conference on Empirical Methods in Natural Language Processing",
    month = dec,
    year = "2023",
    address = "Singapore",
    publisher = "Association for Computational Linguistics",
    url = "https://aclanthology.org/2023.emnlp-main.322",
    doi = "10.18653/v1/2023.emnlp-main.322",
    pages = "5303--5315",
}

@article{yao2024tree,
  title={Tree of thoughts: Deliberate problem solving with large language models},
  author={Yao, Shunyu and Yu, Dian and Zhao, Jeffrey and Shafran, Izhak and Griffiths, Tom and Cao, Yuan and Narasimhan, Karthik},
  journal={Advances in Neural Information Processing Systems},
  volume={36},
  year={2024}
}

@article{feng2023alphazero,
  title={Alphazero-like tree-search can guide large language model decoding and training},
  author={Feng, Xidong and Wan, Ziyu and Wen, Muning and McAleer, Stephen Marcus and Wen, Ying and Zhang, Weinan and Wang, Jun},
  journal={arXiv preprint arXiv:2309.17179},
  year={2023}
}

@inproceedings{liu-etal-2024-knowledge-graph,
    title = "Knowledge Graph-Enhanced Large Language Models via Path Selection",
    author = "Liu, Haochen  and
      Wang, Song  and
      Zhu, Yaochen  and
      Dong, Yushun  and
      Li, Jundong",
    editor = "Ku, Lun-Wei  and
      Martins, Andre  and
      Srikumar, Vivek",
    booktitle = "Findings of the Association for Computational Linguistics: ACL 2024",
    month = aug,
    year = "2024",
    address = "Bangkok, Thailand",
    publisher = "Association for Computational Linguistics",
    url = "https://aclanthology.org/2024.findings-acl.376/",
    doi = "10.18653/v1/2024.findings-acl.376",
    pages = "6311--6321"
}

@inproceedings{Freebase,
author = {Bollacker, Kurt and Evans, Colin and Paritosh, Praveen and Sturge, Tim and Taylor, Jamie},
title = {Freebase: A Collaboratively Created Graph Database for Structuring Human Knowledge},
year = {2008},
isbn = {9781605581026},
publisher = {Association for Computing Machinery},
address = {New York, NY, USA},
url = {https://doi.org/10.1145/1376616.1376746},
doi = {10.1145/1376616.1376746},
booktitle = {Proceedings of the 2008 ACM SIGMOD International Conference on Management of Data},
pages = {1247–1250},
numpages = {4},
location = {Vancouver, Canada},
series = {SIGMOD '08}
}

@inproceedings{zhang-etal-2024-onegen,
    title = "{O}ne{G}en: Efficient One-Pass Unified Generation and Retrieval for {LLM}s",
    author = "Zhang, Jintian  and
      Peng, Cheng  and
      Sun, Mengshu  and
      Chen, Xiang  and
      Liang, Lei  and
      Zhang, Zhiqiang  and
      Zhou, Jun  and
      Chen, Huajun  and
      Zhang, Ningyu",
    editor = "Al-Onaizan, Yaser  and
      Bansal, Mohit  and
      Chen, Yun-Nung",
    booktitle = "Findings of the Association for Computational Linguistics: EMNLP 2024",
    month = nov,
    year = "2024",
    address = "Miami, Florida, USA",
    publisher = "Association for Computational Linguistics",
    url = "https://aclanthology.org/2024.findings-emnlp.237/",
    doi = "10.18653/v1/2024.findings-emnlp.237",
    pages = "4088--4119"
}

@inproceedings{STAGG,
    title = "The Value of Semantic Parse Labeling for Knowledge Base Question Answering",
    author = "Yih, Wen-tau  and
      Richardson, Matthew  and
      Meek, Chris  and
      Chang, Ming-Wei  and
      Suh, Jina",
    booktitle = "Proceedings of the 54th Annual Meeting of the Association for Computational Linguistics (Volume 2: Short Papers)",
    month = aug,
    year = "2016",
    address = "Berlin, Germany",
    publisher = "Association for Computational Linguistics",
    url = "https://aclanthology.org/P16-2033",
    doi = "10.18653/v1/P16-2033",
    pages = "201--206",
}

@inproceedings{CWQ,
    title = "The Web as a Knowledge-Base for Answering Complex Questions",
    author = "Talmor, Alon  and
      Berant, Jonathan",
    booktitle = "Proceedings of the 2018 Conference of the North {A}merican Chapter of the Association for Computational Linguistics: Human Language Technologies, Volume 1 (Long Papers)",
    month = jun,
    year = "2018",
    address = "New Orleans, Louisiana",
    publisher = "Association for Computational Linguistics",
    url = "https://aclanthology.org/N18-1059",
    doi = "10.18653/v1/N18-1059",
    pages = "641--651",
}

@inproceedings{DECAF,
title={Dec{AF}: Joint Decoding of Answers and Logical Forms for Question Answering over Knowledge Bases},
author={Donghan Yu and Sheng Zhang and Patrick Ng and Henghui Zhu and Alexander Hanbo Li and Jun Wang and Yiqun Hu and William Yang Wang and Zhiguo Wang and Bing Xiang},
booktitle={The Eleventh International Conference on Learning Representations },
year={2023},
url={https://openreview.net/forum?id=XHc5zRPxqV9}
}

@inproceedings{sun2019pullnet,
  title     = {PullNet: Open Domain Question Answering with Iterative Retrieval on Knowledge Bases and Text},
  author    = {Sun, Haitian and Bedrax-Weiss, Tania and Cohen, William},
  booktitle = {Proceedings of the 2019 Conference on Empirical Methods in Natural Language Processing and the 9th International Joint Conference on Natural Language Processing (EMNLP-IJCNLP)},
  pages     = {2380--2390},
  year      = {2019}
}

@inproceedings{UniKGQA,
title={Uni{KGQA}: Unified Retrieval and Reasoning for Solving Multi-hop Question Answering Over Knowledge Graph},
author={Jinhao Jiang and Kun Zhou and Xin Zhao and Ji-Rong Wen},
booktitle={The Eleventh International Conference on Learning Representations },
year={2023},
url={https://openreview.net/forum?id=Z63RvyAZ2Vh}
}

@inproceedings{gu-su-2022-arcaneqa,
    title = "{A}rcane{QA}: Dynamic Program Induction and Contextualized Encoding for Knowledge Base Question Answering",
    author = "Gu, Yu  and
      Su, Yu",
    editor = "Calzolari, Nicoletta  and
      Huang, Chu-Ren  and
      Kim, Hansaem  and
      Pustejovsky, James  and
      Wanner, Leo  and
      Choi, Key-Sun  and
      Ryu, Pum-Mo  and
      Chen, Hsin-Hsi  and
      Donatelli, Lucia  and
      Ji, Heng  and
      Kurohashi, Sadao  and
      Paggio, Patrizia  and
      Xue, Nianwen  and
      Kim, Seokhwan  and
      Hahm, Younggyun  and
      He, Zhong  and
      Lee, Tony Kyungil  and
      Santus, Enrico  and
      Bond, Francis  and
      Na, Seung-Hoon",
    booktitle = "Proceedings of the 29th International Conference on Computational Linguistics",
    month = oct,
    year = "2022",
    address = "Gyeongju, Republic of Korea",
    publisher = "International Committee on Computational Linguistics",
    url = "https://aclanthology.org/2022.coling-1.148/",
    pages = "1718--1731"
}

@article{wu2024comparative,
  title={A Comparative Study on Reasoning Patterns of OpenAI's o1 Model},
  author={Wu, Siwei and Peng, Zhongyuan and Du, Xinrun and Zheng, Tuney and Liu, Minghao and Wu, Jialong and Ma, Jiachen and Li, Yizhi and Yang, Jian and Zhou, Wangchunshu and others},
  journal={arXiv preprint arXiv:2410.13639},
  year={2024}
}

@inproceedings{zheng-etal-2024-llamafactory,
    title = "{L}lama{F}actory: Unified Efficient Fine-Tuning of 100+ Language Models",
    author = "Zheng, Yaowei  and
      Zhang, Richong  and
      Zhang, Junhao  and
      Ye, Yanhan  and
      Luo, Zheyan",
    editor = "Cao, Yixin  and
      Feng, Yang  and
      Xiong, Deyi",
    booktitle = "Proceedings of the 62nd Annual Meeting of the Association for Computational Linguistics (Volume 3: System Demonstrations)",
    month = aug,
    year = "2024",
    address = "Bangkok, Thailand",
    publisher = "Association for Computational Linguistics",
    url = "https://aclanthology.org/2024.acl-demos.38/",
    doi = "10.18653/v1/2024.acl-demos.38",
    pages = "400--410"
}

@inproceedings{rajbhandari2020zero,
  title={Zero: Memory optimizations toward training trillion parameter models},
  author={Rajbhandari, Samyam and Rasley, Jeff and Ruwase, Olatunji and He, Yuxiong},
  booktitle={Proceedings of the International Conference for High Performance Computing, Networking, Storage and Analysis},
  year={2020},
  url="https://dl.acm.org/doi/10.5555/3433701.3433727"
}

@article{dubey2024llama,
  title={The llama 3 herd of models},
  author={Dubey, Abhimanyu and Jauhri, Abhinav and Pandey, Abhinav and Kadian, Abhishek and Al-Dahle, Ahmad and Letman, Aiesha and Mathur, Akhil and Schelten, Alan and Yang, Amy and Fan, Angela and others},
  journal={arXiv preprint arXiv:2407.21783},
  year={2024}
}

@article{zhang2024alter,
  title={Alter: Augmentation for large-table-based reasoning},
  author={Zhang, Han and Ma, Yuheng and Yang, Hanfang},
  journal={arXiv preprint arXiv:2407.03061},
  year={2024}
}

@article{codeKG,
author = {Bi, Zhen and Chen, Jing and Jiang, Yinuo and Xiong, Feiyu and Guo, Wei and Chen, Huajun and Zhang, Ningyu},
title = {CodeKGC: Code Language Model for Generative Knowledge Graph Construction},
year = {2024},
issue_date = {March 2024},
publisher = {Association for Computing Machinery},
address = {New York, NY, USA},
volume = {23},
number = {3},
issn = {2375-4699},
url = {https://doi.org/10.1145/3641850},
doi = {10.1145/3641850},
journal = {ACM Trans. Asian Low-Resour. Lang. Inf. Process.},
month = mar,
articleno = {45},
numpages = {16}
}

@article{jiang2024hykge,
  title={HyKGE: A Hypothesis Knowledge Graph Enhanced Framework for Accurate and Reliable Medical LLMs Responses},
  author={Jiang, Xinke and Zhang, Ruizhe and Xu, Yongxin and Qiu, Rihong and Fang, Yue and Wang, Zhiyuan and Tang, Jinyi and Ding, Hongxin and Chu, Xu and Zhao, Junfeng and others},
  journal={arXiv preprint arXiv:2312.15883},
  year={2024}
}

@inproceedings{zhang2018variational,
  title     = {Variational reasoning for question answering with knowledge graph},
  author    = {Zhang, Yuyu and Dai, Hanjun and Kozareva, Zornitsa and Smola, Alexander and Song, Le},
  booktitle = {Proceedings of the AAAI conference on artificial intelligence},
  volume    = {32},
  number    = {1},
  year      = {2018}
}

@article{xie2023self,
  title={Self-evaluation guided beam search for reasoning},
  author={Xie, Yuxi and Kawaguchi, Kenji and Zhao, Yiran and Zhao, James Xu and Kan, Min-Yen and He, Junxian and Xie, Michael},
  journal={Advances in Neural Information Processing Systems},
  volume={36},
  pages={41618--41650},
  year={2023}
}

@article{guan2025rstar,
  title={rStar-Math: Small LLMs Can Master Math Reasoning with Self-Evolved Deep Thinking},
  author={Guan, Xinyu and Zhang, Li Lyna and Liu, Yifei and Shang, Ning and Sun, Youran and Zhu, Yi and Yang, Fan and Yang, Mao},
  journal={arXiv preprint arXiv:2501.04519},
  year={2025}
}

@misc{bge_embedding,
      title={C-Pack: Packaged Resources To Advance General Chinese Embedding}, 
      author={Shitao Xiao and Zheng Liu and Peitian Zhang and Niklas Muennighoff},
      year={2023},
      eprint={2309.07597},
      archivePrefix={arXiv},
      primaryClass={cs.CL}
}

@inproceedings{li-etal-2022-c3kg,
    title = "{C}$^3${KG}: A {C}hinese Commonsense Conversation Knowledge Graph",
    author = "Li, Dawei  and
      Li, Yanran  and
      Zhang, Jiayi  and
      Li, Ke  and
      Wei, Chen  and
      Cui, Jianwei  and
      Wang, Bin",
    editor = "Muresan, Smaranda  and
      Nakov, Preslav  and
      Villavicencio, Aline",
    booktitle = "Findings of the Association for Computational Linguistics: ACL 2022",
    month = may,
    year = "2022",
    address = "Dublin, Ireland",
    publisher = "Association for Computational Linguistics",
    url = "https://aclanthology.org/2022.findings-acl.107/",
    doi = "10.18653/v1/2022.findings-acl.107",
    pages = "1369--1383"
}

@inproceedings{li-etal-2020-efficient,
    title = "Efficient One-Pass End-to-End Entity Linking for Questions",
    author = "Li, Belinda Z.  and
      Min, Sewon  and
      Iyer, Srinivasan  and
      Mehdad, Yashar  and
      Yih, Wen-tau",
    editor = "Webber, Bonnie  and
      Cohn, Trevor  and
      He, Yulan  and
      Liu, Yang",
    booktitle = "Proceedings of the 2020 Conference on Empirical Methods in Natural Language Processing (EMNLP)",
    month = nov,
    year = "2020",
    address = "Online",
    publisher = "Association for Computational Linguistics",
    url = "https://aclanthology.org/2020.emnlp-main.522/",
    doi = "10.18653/v1/2020.emnlp-main.522",
    pages = "6433--6441"
}

@inproceedings{ye-etal-2022-rng,
    title = "{RNG}-{KBQA}: Generation Augmented Iterative Ranking for Knowledge Base Question Answering",
    author = "Ye, Xi  and
      Yavuz, Semih  and
      Hashimoto, Kazuma  and
      Zhou, Yingbo  and
      Xiong, Caiming",
    editor = "Muresan, Smaranda  and
      Nakov, Preslav  and
      Villavicencio, Aline",
    booktitle = "Proceedings of the 60th Annual Meeting of the Association for Computational Linguistics (Volume 1: Long Papers)",
    month = may,
    year = "2022",
    address = "Dublin, Ireland",
    publisher = "Association for Computational Linguistics",
    url = "https://aclanthology.org/2022.acl-long.417/",
    doi = "10.18653/v1/2022.acl-long.417",
    pages = "6032--6043"
}
\end{document}